\documentclass[sigconf]{acmart}
\usepackage{url}
\usepackage{kantlipsum, lipsum}
\usepackage{dsfont}
\usepackage{subfigure}
\usepackage{subcaption}
\usepackage{bbm}
\usepackage{wrapfig}
\usepackage{enumitem}
\usepackage{microtype}
\usepackage[dvipsnames]{xcolor}
\usepackage{amsmath}
\usepackage{mathtools}
\usepackage{hyperref}
\usepackage[capitalize,noabbrev,nameinlink]{cleveref}

\usepackage{etoolbox}

\crefname{appendix}{Appendix}{Appendices}
\Crefname{appendix}{Appendix}{Appendices}

\makeatletter
\pretocmd\appendix{%
  \crefalias{section}{appendix}%
}{}{}
\makeatother

\usepackage{booktabs}
\usepackage{annotate_equations}
\usepackage[export]{adjustbox}

\usepackage{algorithm}
\usepackage[noend]{algpseudocode}

\usepackage[most,skins,theorems]{tcolorbox}
\tcbset{
  aibox/.style={
    width=\linewidth,
    top=8pt,
    bottom=4pt,
    colback=blue!6!white,
    colframe=black,
    colbacktitle=black,
    enhanced,
    center,
    attach boxed title to top left={yshift=-0.1in,xshift=0.15in},
    boxed title style={boxrule=0pt,colframe=white,},
  }
}
\newtcolorbox{AIbox}[2][]{aibox,title=#2,#1}

\AtBeginDocument{%
  }

\copyrightyear{2026}
\acmYear{2026}
\setcopyright{cc}
\setcctype{by}
\acmConference[KDD 2026] {Proceedings of the 32nd ACM SIGKDD Conference on Knowledge Discovery and Data Mining V.2}{August 9--13, 2026}{Jeju Island, Republic of Korea.}
\acmBooktitle{Proceedings of the 32nd ACM SIGKDD Conference on Knowledge Discovery and Data Mining V.2 (KDD 2026), August 9--13, 2026, Jeju Island, Republic of Korea}
\acmISBN{979-8-4007-2259-2/2026/08}
\acmDOI{10.1145/3770855.3817976}

\begin{document}

\title{Understanding Generative Recommendation with Semantic IDs from a Model-scaling View}

\author{Jingzhe Liu}
\email{liujin33@msu.edu}
\affiliation{%
  \institution{Michigan State University}
  \city{East Lansing}
  \state{Michigan}
  \country{USA}
}

\author{Liam Collins}
\email{lcollins2@snapchat.com}
\affiliation{%
  \institution{Snap Inc.}
  \city{Bellevue}
  \state{Washington}
  \country{USA}
  }

\author{Jiliang Tang}
\email{tangjili@msu.edu}
\affiliation{%
  \institution{Michigan State University}
  \city{East Lansing}
  \state{Michigan}
  \country{USA}
}

\author{Tong Zhao}
\email{tong@snap.com}
\affiliation{%
  \institution{Snap Inc.}
  \city{Bellevue}
  \state{Washington}
  \country{USA}
  }

\author{Neil Shah}
\email{nshah@snap.com}
\affiliation{%
  \institution{Snap Inc.}
  \city{Bellevue}
  \state{Washington}
  \country{USA}
  }

\author{Mingxuan Ju}
\email{mju@snap.com}
\affiliation{%
  \institution{Snap Inc.}
  \city{Bellevue}
  \state{Washington}
  \country{USA}
  }

\renewcommand{\shortauthors}{Jingzhe Liu et al.}

\begin{abstract}
Recent advancements in generative models have allowed the emergence of a promising paradigm for recommender systems (RS), known as Generative Recommendation (GR), which tries to unify rich item semantics and collaborative filtering signals. 
One popular GR paradigm is to use semantic IDs (SIDs), which are discrete codes quantized from the embeddings of modality encoders (e.g. large language or vision models), to represent items in an autoregressive user interaction sequence modeling setup (henceforth, \textit{SID-based GR}).
While generative models in other domains exhibit well-established scaling laws, our work reveals that SID-based GR shows significant bottlenecks while scaling up the model; in particular, the performance of SID-based GR quickly saturates as we enlarge each component -- the modality encoder, the quantization tokenizer, and the RS itself. 
In this work, we identify \textit{the limited capacity of SIDs to encode item semantic information as one of the fundamental bottlenecks}. 
Motivated by this observation, as an initial effort to obtain GR models with better scaling behaviors, we revisit another GR paradigm that directly uses textual representations of items (henceforth, \textit{Text-based GR}).
Our experiments show that Text-based GR paradigm has superior model scaling properties and achieves up to \textbf{20\%} improvement over the best achievable performance of SID-based GR through scaling. 
We also challenge the prevailing belief that LLMs struggle to capture collaborative filtering information, showing that LLMs' ability to model user–item interactions improves as LLMs scale up. 
Our analyses across model sizes from 44M to 14B parameters underscore the intrinsic scaling limits of SID-based GR and position Text-based GR as a promising path toward foundation models for GR.
\end{abstract}




\maketitle

\section{Introduction}\label{sec:intro}


Recommender systems (RS) — such as those for products~\citep{wang2021dcn,schafer1999recommender}, videos~\citep{gomez2015netflix,van2013deep}, and friends~\citep{sankar2021graph, ju2025learning} — play a pivotal role in tailoring personalized experiences for millions of users and driving enhanced engagement with online platforms. 
Conventionally, RS follows a two-stage pipeline: \textit{retrieval-and-ranking}~\citep{huang2020embedding,weller2025theoretical}. 
During the retrieval stage, a large candidate item corpus is first trimmed down to a subset of items, leveraging computationally efficient heuristics or lightweight neural models, which are quickly processed to filter for broad relevance.
This is followed by the ranking stage, where fine-grained user preferences are produced, utilizing complex features that capture intricate patterns, enabling the system to precisely score and order the candidate items~\citep{wang2021dcn,zhou2018deep}.

\begin{figure*}[t]
\begin{center}
\includegraphics[width=0.8\textwidth]{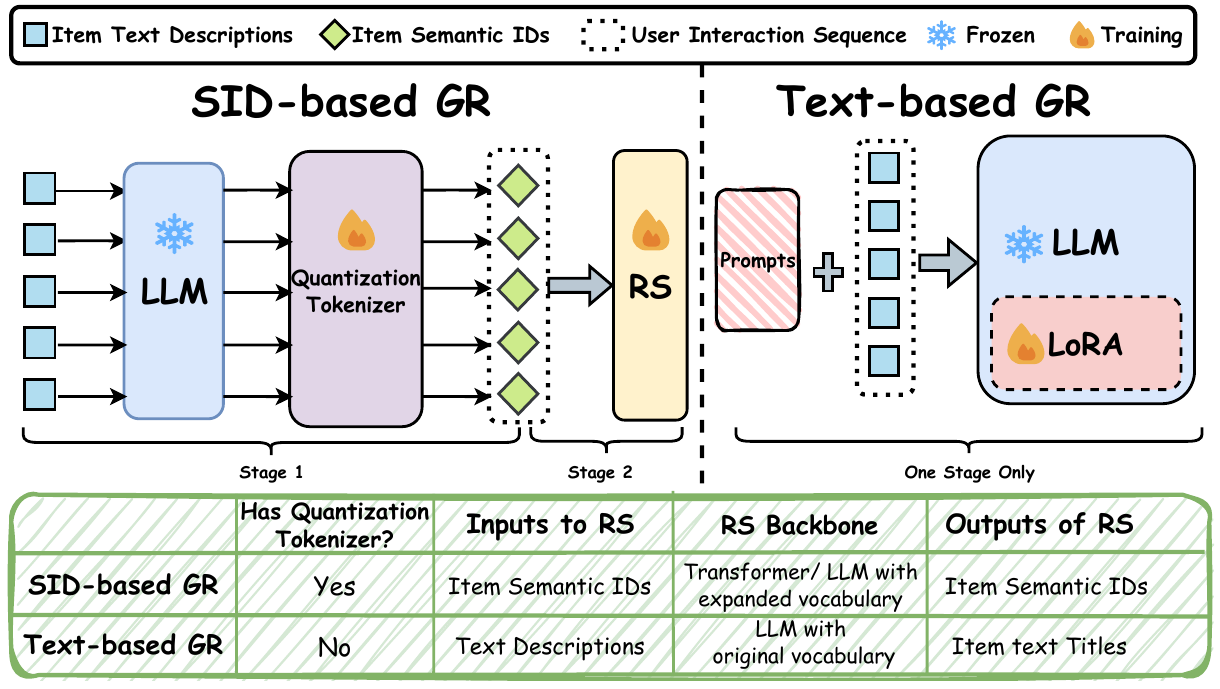}
\end{center}

\caption{Two GR paradigms we investigate in this paper. SID-based GR first transforms the item textual descriptions into semantic IDs and then trains a transformer or LLM with expanded vocabulary to predict the SIDs of the next item, while Text-based GR directly takes in the texts and outputs the title of the next item.}
\label{fig:paradigms}
\end{figure*}

Recently, generative recommendation (GR) is emerging as a transformative paradigm of RS, attracting significant attention~\citep{wang2023generative, TIGER, wang2024eager} and seeing wide adoption in real-world applications~\citep{trillion-param-genrec, OneRec, Tandon2025GeminiYouTube}. Unlike conventional methods, GR capitalizes on recent advancements in generative models~\citep{guo2025deepseek, qwen3} and directly generates recommendations in an \textit{end-to-end} manner. 
It either directly generates text descriptions of recommended items~\citep{P5, IDGenRec, hua2023index, bao2023tallrec} or uses pre-trained models to extract rich semantic representations that encode broad, open-world knowledge~\citep{yuan2023go, ren2024representation, LIGER}.
This emerging approach not only provides additional knowledge from pre-trained models but also avoids the complexity and potential errors associated with multi-stage systems.

Specifically, in GR, semantic IDs (SIDs)~\citep{TIGER,LIGER,OneRec, luo2024qarm, zheng2025enhancing} are a widely adopted mechanism for integrating existing powerful foundation models with RS. 
As \cref{fig:paradigms} illustrates, this paradigm consists of two main stages. 
First, a modality encoder (e.g., LLM or VLM) and a quantization tokenizer (e.g., RQ-VAE~\citep{RQ22}, VQ-VAE~\citep{esser2021taming}, or Residual K-Means~\citep{OneRec}) convert item content features (such as image or text) into SIDs. 
Subsequently, a sequential recommender is trained to autoregressively predict the SIDs corresponding to future user interactions based on the history of previously consumed SIDs~\citep{TIGER}.
GR using SIDs aims to unify semantic knowledge from pretrained foundation models with collaborative filtering~(CF) signals from user interactions. 
The overlap between item SIDs inherently captures item content information to leverage priors of semantic similarity, while next-item prediction supervision enables the model to learn user-behavior patterns, which covers the two pivotal types of knowledge for an effective sequential recommender. 

Notably, generative models in other domains have been demonstrated to exhibit scaling behaviors, where the performance improves predictably as a function of increased model size, dataset size, and/or computational budget~\citep{kaplan2020scaling,mixed-modal-scaling,VAR,retrieval_scaling}.
Such observation of scaling laws is not only an academic curiosity, but is critical in informing how to efficiently scale next-generation models~\citep{chung2024scaling,GPT4,qwen3}, providing a framework for assessing different model architectures and enabling researchers to identify model classes with more favorable properties. 


In this work, we investigate the model scaling behaviors of the SID-based GR paradigm. To do so, we experiment with the impact of model scale in both SID-generation (encoder, quantizer) and autoregressive decoding (sequential recommender) using LLMs of different sizes. 
Importantly, we empirically discover that the recommendation performance saturates very fast as we scale up the GR pipeline from multiple axes.
Such counterintuitive finding stands in stark contrast to the well-established scaling laws observed in other generative domains, suggesting that the benefits of scale do not transfer straightforwardly to SID-based GR.
This phenomenon suggests that the SID-based paradigm may contain fundamental bottlenecks, preventing the straightforward application of scaling laws and hence underscoring the need for a more nuanced understanding of how scaling law interacts with GR. 
Our work aims to understand the following fundamental research questions:
\begin{center}
    \textit{\textbf{What are key bottlenecks preventing SID-based GR models from scaling up? Are there other GR paradigms that can overcome them and exhibit better scaling behaviors? }}
\end{center}


Existing works in GR strengthen the performance by either incorporating external knowledge such as additional CF signals~\citep{A-LLMRec,LLaRA} or richer item semantic information from LLMs~\citep{LIGER, COBRA}.
Beyond these approaches, little attention has been given to whether GR models can be improved simply by scaling up their parameters to better capture item and user information.
Drawing on the experience of LLMs, we argue that scaling core architectural components~\citep{kaplan2020scaling,GPT3} represents a promising direction toward more performant GR models and, ultimately, recommendation foundation models.
We delve into this direction by conducting a comprehensive analysis on the effects of scaling up GR models and summarize our contributions as follows:
\begin{itemize}[leftmargin=*,topsep=0pt]
    \item 
We show that scaling sequential recommenders~(RS) in SID-based GR models results in early performance saturation~(vanilla transformer) or marginal gains~(LLM), while neither the modality encoder nor the quantization tokenizer exhibits scaling behaviors. 
Our analysis further reveals that the fundamental bottleneck of SID-based GR models lies in their limited capacity to capture item content information. 
In particular, the SID itself constitutes the key constraint, preventing the knowledge embedded in the powerful modality encoders~(i.e., LLMs) from being effectively transferred into the downstream recommender.
    
    \item 
    Built atop our findings above, we revisit the GR paradigm of directly using textual item representations (dubbed Text-based GR as in \cref{fig:paradigms}) and show it has \textbf{better scaling behaviors}. Moreover, we challenge the prevailing view that it lacks the ability to capture CF signals. 
    Instead, we demonstrate it is capable of modeling CF information, with this capability improving as model size increases. 
    Consequently, the benefits of incorporating external CF embeddings decrease as the backbone scales up. 
    \item 
    Overall, we conclude that the Text-based GR paradigm exhibits superior scaling properties compared to SID-based GR models. Its performance improves consistently with model size, showing no signs of saturation. Furthermore, when being scaled up, Text-based GR quickly surpasses SID-based GR and achieves up to a 20\% improvement using the same training data.
\end{itemize}

\textbf{Clarification}: We note that \textit{this study does not aim to achieve state-of-the-art performance, nor to claim that one paradigm is universally superior to the other under all circumstances}. Rather, our goal is to provide a thorough understanding of their respective advantages and limitations through the model-scaling behaviors analysis, thereby paving the way toward building effective GR models.

\section{Scaling Law for Generative Recommendation}\label{sec:setup}


Given an effective GR model, it should well capture two essential types of information: \textbf{\textit{collaborative filtering signals (CF)}}  and \textbf{\textit{semantic information (SI)}}~\citep{zhou2020s3,Survey-LLM4Rec, sheng2024language}. 
The former refers to how well the model learns the interaction histories, a cornerstone of conventional RS~\citep{SASRec,sun2019bert4rec}; and the latter refers to how well the model learns the semantic content (e.g., text or image) of items, which recently has shown to be highly effective for recommendation~\citep{TIGER,Survey-LLM4Rec,P5}.

Hence, we posit that a good (in terms of scaling behaviors) GR model \textit{should exhibit improved ability to capture both CF as well as SI as the model progressively scales up}.
To quantitatively examine the capability of GR models in modeling CF and SI, we begin by formulating the scaling law equation under the classical risk decomposition principle~\citep{hoffmann2022training}.
Specifically, we regard CF and SI as two distinct modalities~\citep{LLaRA} and propose the scaling equation for GR models (\cref{eq:basic_scaling_law}), following previous works of scaling laws of multimodal models~\citep{mixed-modal-scaling,vocab_scaling}.
Since we focus on model scaling, \cref{eq:basic_scaling_law} assumes a fixed amount of training data and characterizes how the overall error~($\mathcal{L}$) varies with the number of model parameters allocated to SI learning~($N_{\text{SI}}$) and CF learning~($N_{\text{CF}}$), where the exact forms of $N_{\text{SI}}$ and $N_{\text{CF}}$ depend on the specific model~(they can share same terms since some modules learn both CF and SI).
The overall loss~($\mathcal{L}$) consists of three additive terms: the minimal achievable error~($E$), the semantic information error~($\tfrac{A}{N_{\text{SI}}^b}$), and the collaborative filtering error~($\tfrac{B}{N_{\text{CF}}^a}$), where $E$, $A$, $B$, $a$, and $b$ are positive and empirically determined parameters, formulated as:

\begin{figure}[h]
\vspace{1em}
\begin{equation}
\label{eq:basic_scaling_law}
\mathcal{L}
    \left(\eqnmarkbox[NavyBlue]{Ni}{N_{\text{SI}}}, 
    \eqnmarkbox[OliveGreen]{Nu}{N_{\text{CF}}}\right) 
    = \eqnmarkbox[Plum]{E}{E}  + 
    \eqnmarkbox[BurntOrange]{conv}{\frac{A}{N_{\text{SI}}^a}}   
    + 
    \eqnmarkbox[Emerald]{func}{\frac{B}{N_{\text{CF}}^{b}}}.
\end{equation}
\annotate[yshift=1em]{above,left}{Ni}{Params for SI learning}
\annotate[yshift=-1em]{below,left}{Nu}{Params for CF learning}
\annotate[yshift=-2.5em]{below,left}{E}{Minimal Achievable Error}
\annotate[yshift=1em]{above,right}{func}{CF-related Error}
\annotate[yshift=-0.5em]{below,right}{conv}{SI-related Error}

\vspace{1.5em}
\end{figure}

Given the retrieval nature of GR and RS, we follow \citet{retrieval_scaling} and adopt the Miss Rate~($\mathrm{MR@}k$) as the metric for the overall error~$\mathcal{L}$, where $\mathcal{L}=\mathrm{MR@}k = 1 - \mathrm{Recall@}k$ and $k$ is an integer.
To align with the previous works, we report $\mathrm{Recall@}k$ to represent model performance, which allows \cref{eq:basic_scaling_law} to be rewritten as:

\begin{equation}
\label{eq:empirical_scaling_law}
\mathrm{Recall@}k
= R_0 -
\frac{A}{N_{\text{SI}}^a}
- \frac{B}{N_{\text{CF}}^b},
\end{equation}
where $R_0$ denotes the maximal achievable performance ($0 < R_0 < 1$) and is also learned empirically. All subsequent empirical analyses are grounded in \cref{eq:empirical_scaling_law}.

\textbf{Datasets.} We conduct our experiments on the Amazon Review datasets~\citep{he2016ups,BLAIR}. 
We use three subdatasets: {\small\textsf{Beauty}}, {\small\textsf{Sports and Outdoors}}, and {\small\textsf{Toys and Games}}. 
Since we focus on model scaling in this research, we fix the training and evaluation data amount for the experiments in this paper.
We follow the same train/valid/test dataset splits and pre-processing as \cite{ju2025generative} by default.
The dataset and implementation details can be found in \cref{app:dataset_details}. 

\section{Investigating the Model Scaling Behaviors of SID-based GR}\label{sec:scaling_codewords}

\textbf{Setup.} To investigate the scaling properties of the SID-based GR, we use the cornerstone framework which is widely shared by transformer-RS methods~(e.g., TIGER~\citep{TIGER}) and LLM-RS methods~(e.g., MiniOneRec~\citep{minionerec}).
As shown in \cref{fig:paradigms}, its training process involves two main stages: (1) a modality encoder (LLM in this case) encodes item descriptions into semantic embeddings, which are then quantized into discrete SIDs, and (2) a sequential recommender is trained on user SID sequences to predict the next item's SID.
During inference, we generate next-item SIDs using beam search.
Specifically, our experiments leverage the open-source framework GRID~\citep{ju2025generative}. 
We first generate the semantic IDs with its codes and then input them to transformer RS following TIGER~\citep{TIGER} and to LLM RS following MiniOneRec~\citep{minionerec} respectively.
To align the learning strategy and ensure fair comparisons, all RS models are trained via supervised tuning.

As illustrated in \cref{fig:paradigms}, SID-based GR contains three overarching components: the pre-trained frozen LLM modality encoder~($N_{\text{LLM}}$), the trainable quantization tokenizer~($N_{\text{QT}}$), and the trainable downstream recommender~($N_{\text{RS}}$).
Since the LLM encoder and tokenizer process only semantic information, we set $N_{\text{CF}} = N_{\text{RS}}$ and $N_{\text{SI}} = N_{\text{RS}} + \gamma_1 N_{\text{LLM}} + \gamma_2 N_{\text{QT}}$, where $0\leq \gamma_1, \gamma_2 \leq 1$ are effective parameter coefficients of the LLM encoder and quantization tokenizer~(since they are frozen during the training stage of RS), respectively.
Hence, \cref{eq:empirical_scaling_law} can be reformulated as:
\begin{figure}[h]
\vspace{0.7em}
\begin{equation}
\label{eq:codewords_scaling_law}
\mathrm{Recall@}k
    = R_0  -
    \eqnmarkbox[BurntOrange]{item}{\frac{A}{(N_{\text{RS}}+\gamma_1 N_{\text{LLM}}+\gamma_2 N_{\text{QT}})^a}}
    -
    \eqnmarkbox[Emerald]{user}{\frac{B}{N_{\text{RS}}^{b}}}.
\end{equation}

\annotate[yshift=1em]{above,left}{item}{SI Error}
\annotate[yshift=1em]{above,right}{user}{CF Error}

\vspace{-1em}
\end{figure}

In the following, we individually examine how model performance varies with $N_{\text{RS}}$, $N_{\text{LLM}}$, and $N_{\text{QT}}$.
In \cref{subsec:code_RS_scaling}, we first show that the sequential recommender quickly reaches a performance plateau as we progressively scale up $N_{\text{RS}}$, indicating that its scaling capabilities are limited. 
Then in \cref{subsec:code_LLM_scaling} and \cref{subsec:code_VQ_scaling}, we demonstrate that neither the LLM encoder $N_{\text{LLM}}$ nor the quantization tokenizer $N_{\text{QT}}$ exhibits meaningful scaling behavior on its own within this specific framework. 
Combining these takeaways, in \cref{subsec:ablation on codewords+RS}, our studies reveal that the primary bottleneck to scaling up SID-based GR is its limited ability of encoding semantic information, which is a key component of the model's success.
These observations highlight a fundamental limitation of the model scaling behaviors of SID-based paradigm.

\subsection{Scaling Up the RS Module~($N_{\text{RS}}$) Quickly Saturates the Performance}\label{subsec:code_RS_scaling}
To start with, we examine how performance scales with the sequential recommender.
To be comprehensive, we investigate two different RS backbones: the encoder-decoder transformer of TIGER and the decoder-only LLM of MiniOneRec. 
Specifically, we enlarge the recommender model size~($N_{\text{RS}}$ in \cref{eq:codewords_scaling_law}) and plot how the performance changes for each dataset.
In the experiments, we vary the size of transformer RS from 0.3M
to 3.1B, and the size of LLM RS varies from 0.6B to 14B (we use the Qwen3 models).
We utilize Flan-T5-Large~\citep{chung2024scaling} as the LLM encoder (i.e., $N_{\text{LLM}}= 783$M) and explore RQ-VAE~\citep{RQ22} as QT to generate SIDs with codebooks of shape $3\times256$~(detailed in \cref{subsec:app_experiment_sid}), following the community convention~\citep{TIGER,ju2025generative}.
More details of the experiments can be found in \cref{app:experiment details}.

\begin{figure*}[t]

\begin{center}

\includegraphics[width=0.9\textwidth]{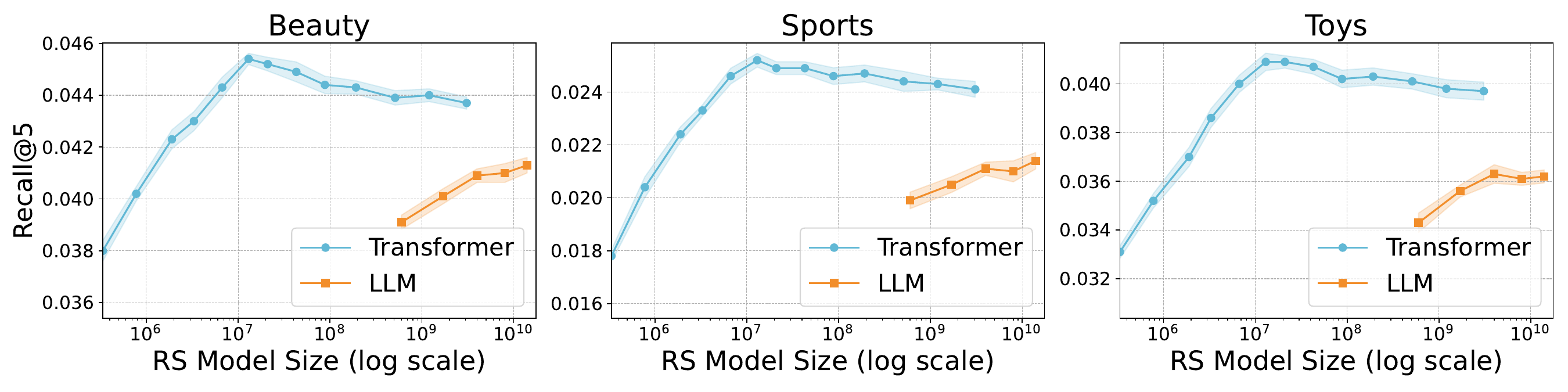}
\end{center}
\caption{The recommendation performance with varying RS model sizes ($N_{\text{RS}}$). The size of transformer RS varies from 0.3M to 3.1B, and the size of LLM RS varies from 0.6B to 14B (Qwen3-series models are used). The performance of transformer RS quickly saturates as $N_{\text{RS}}$ scales up to $10^7$ parameters. Meanwhile, scaling LLM RS provides only marginal gains and consistently underperforms the transformer RS. The shadow regions around the lines indicate error bars from three repeated runs.}
\label{fig:code RS scaling}

\end{figure*}

\begin{figure*}[t]
\begin{center}
\includegraphics[width=0.9\textwidth]{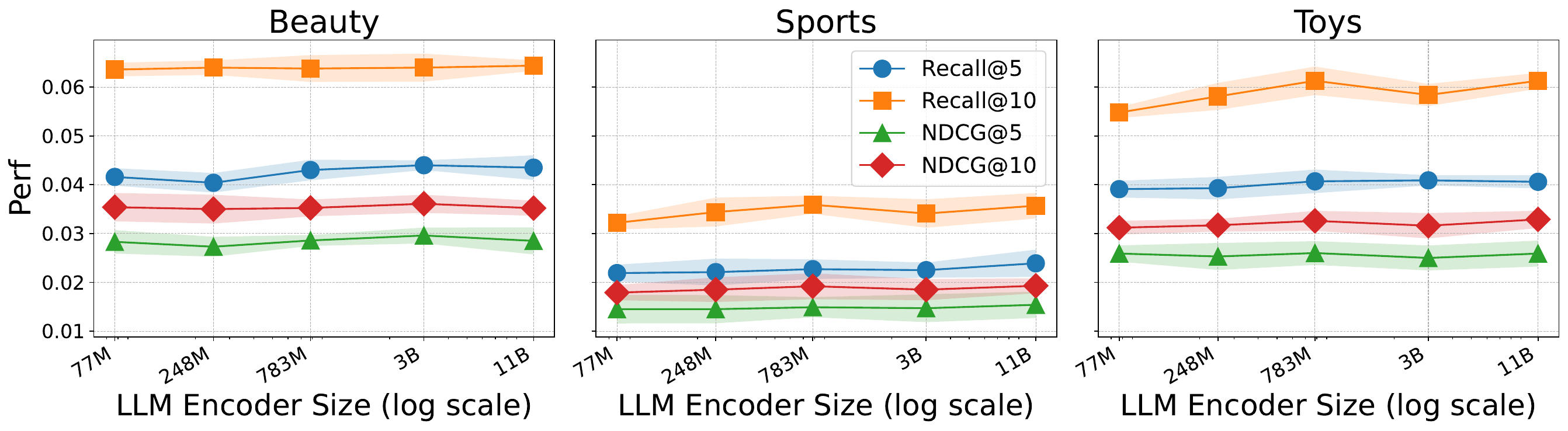}
\end{center}
\caption{The recommendation performance with varying LLM modality encoder sizes ($N_{\text{LLM}}$). Little to no effective scaling behaviors are observed. The shadow regions around the lines indicate error bars from three repeated runs.}
\label{fig:code LLM scaling}

\end{figure*}

\textbf{Results.} The scaling behaviors are shown in \cref{fig:code RS scaling}~(please refer to \cref{app:equation3_fit} for the results of fitting scaling equation). 
On all the three datasets, the performance of transformer RS keeps increasing when $N_{\text{RS}}$ is small. This means that the model scaling already reaches its upper limit.
However, when $N_{\text{RS}}$ is larger than 13M, keeping enlarging it does not bring any gains.
On the other hand, scaling up the LLM RS always results in marginal performance gains, and its performance is inferior to the transformer RS.
Hence, we make our first observation:

\textbf{\textit{Observation 1}: The transformer RS module exhibits scaling behaviors at small scales (i.e., $N_{\text{RS}} \leq 10^7$), but the performance saturates fast under fixed-data regime. The LLM RS has marginal scaling gains and inferior performance. Both RS modules exhibit model scaling bottlenecks.
}

\subsection{Scaling Up the Modality Encoder~($N_{\text{LLM}}$) Brings Little Performance Benefits}\label{subsec:code_LLM_scaling}

Based on observations in \cref{subsec:code_RS_scaling}, a key question is \textit{whether we can overcome the scaling saturation of $N_{\text{RS}}$ by increasing $N_{\text{LLM}}$ or $N_{\text{QT}}$}?
In other words, do larger encoders or tokenizers produce more informative item SIDs, which in turn improve the scaling properties of RS? 
Our scaling law formulations imply this is true if the coefficients $\gamma_1, \gamma_2 > 0$ in \cref{eq:codewords_scaling_law}.

We first examine the scaling behavior of the LLM encoder, which transforms item text descriptions into semantic embeddings.
To scale up the encoder, we employ the Flan-T5 series~\citep{chung2024scaling} with varying sizes ranging from 77M to 11B parameters.
Following the optimal setup in the previous section, we fix the number of codebooks in QT at 3, the size of each code book at 256, and $N_{\text{RS}}$ at $\sim$13M. The isolation of variables allows us to pinpoint the specific contribution of $N_{\text{LLM}}$ or $N_{\text{QT}}$. 
We hypothesize that a larger, more capable LLM should generate richer, more semantically meaningful SIDs, which in turn could enhance the recommendation system's performance.

\textbf{Results.}
Our analysis, based on metrics like Recall@5, Recall@10, NDCG@5, and NDCG@10, reveals a surprising and critical finding: scaling the LLM encoder size provides negligible to no performance improvement across all three datasets. 
As seen in our scaling curves, the performance plateaued almost immediately, with no significant gains from using an 11B parameter model over a much smaller 77M parameter model. 
This suggests that the semantic embedding quality generated by the LLM doesn't act as a bottleneck for the system's performance. 
The item SIDs produced by even the smallest LLM seem to be sufficiently rich for the task, and adding more parameters doesn't significantly enhance their informativeness. 
We therefore summarize the following observation:

\textbf{\textit{Observation 2}: The LLM encoder (i.e., $N_{\text{LLM}}$) in the SID-based GR paradigm shows little to no scaling behaviors~(i.e., $\gamma_1 \approx 0$ in \cref{eq:codewords_scaling_law}).
}

\subsection{Scaling Up the Quantization Tokenizer~($N_{\mathrm{QT}}$) Quickly Saturates the Performance As Well}\label{subsec:code_VQ_scaling}

Having shown that scaling up LLM encoders ($N_{\text{LLM}}$) does not overcome performance saturation, we next investigate if enlarging the quantized tokenizer~($N_{\text{QT}}$) is more effective. The RQ-VAE tokenizer architecture~\citep{RQ21,RQ22} provides two primary scaling dimensions~(detailed in \cref{subsec:app_experiment_sid}): (1) increasing the number of codebooks and (2) increasing the codebook size (cardinality).
In the following experiments, we evaluate both approaches. Specifically, we fix the codebook size at 256 when varying the number of codebooks, and fix the number of codebooks at 3 when varying the size of each codebook.

Note that modifying the tokenizer's configuration also changes the total SID vocabulary, which directly impacts the embedding table size of the subsequent RS transformer. 
To account for this and avoid suboptimal scaling behaviors~\citep{vocab_scaling}, we scale the RS module in tandem with the tokenizer. 
Specifically, we vary the number of codebooks from 2 to 16 (while fixing codebook size at 256) and the codebook size from 32 to 1024 (while fixing the number of codebooks at 3). 
Concurrently, we scale the non-embedding parameters of the RS module from 3M to 21M. 
To ensure the input features are sufficiently expressive, we use the largest LLM encoder (i.e., 11B).

\begin{figure*}[t]
\begin{center}

\includegraphics[width=0.9\textwidth]{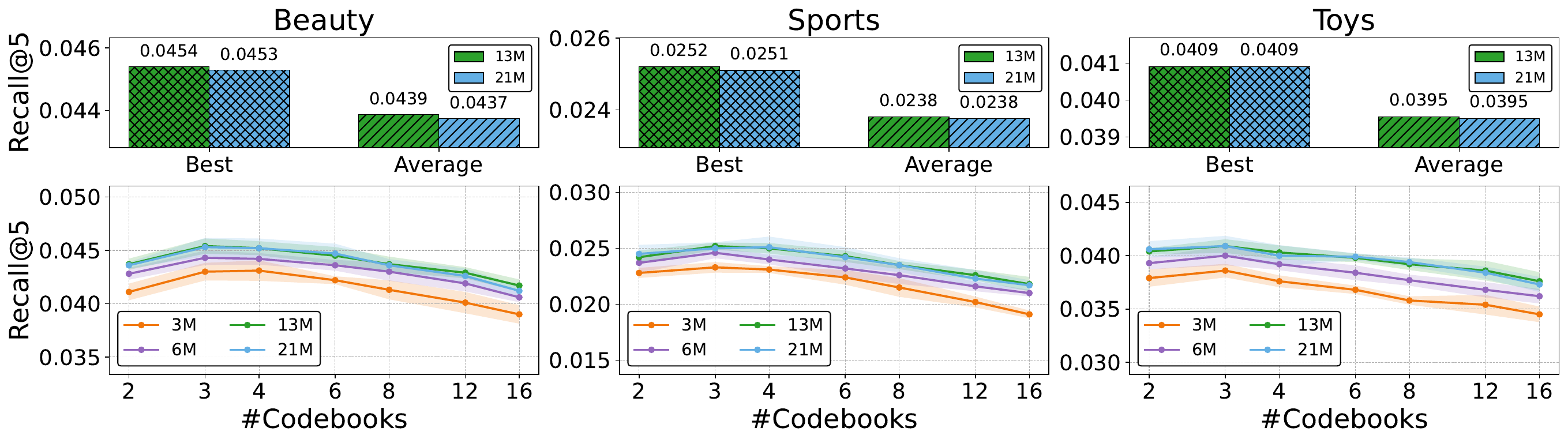}
\end{center}
\vspace{-1em}
\caption{\textit{Lower}: Scaling behaviors of quantization tokenizer when varying the number of codebooks. \textit{Upper}: Comparison of performances between RS modules of 13M and 21M parameters. Overall, increasing the number of codebooks does not overcome the scaling saturation. The shadow regions around the lines indicate error bars from three repeated runs.}

\label{fig:codebook scaling}
\end{figure*}

\begin{figure*}[t]
\begin{center}
\includegraphics[width=0.9\textwidth]{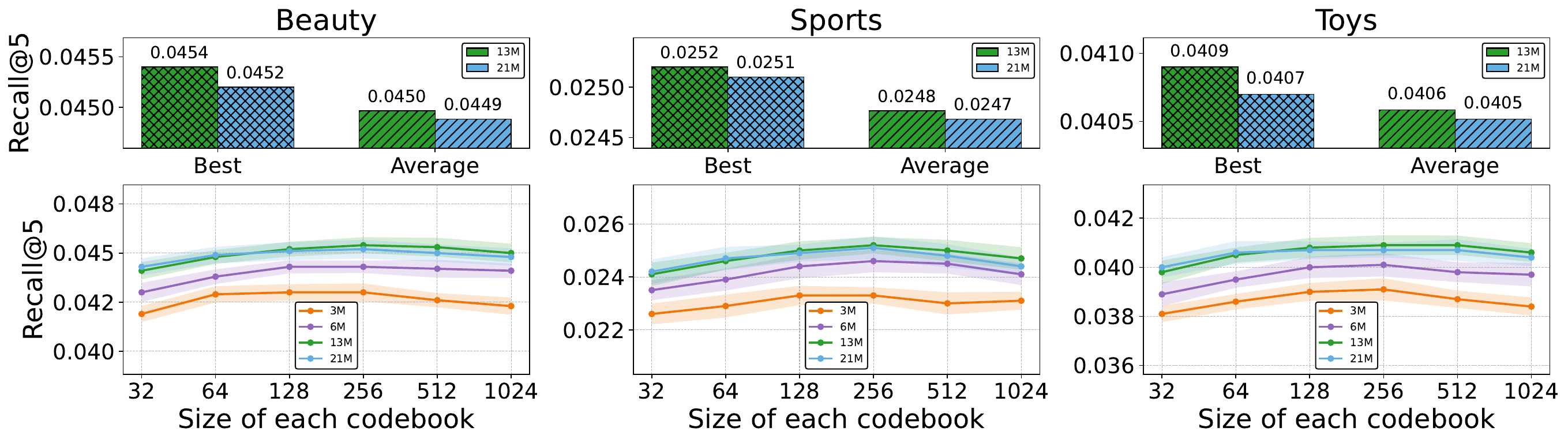}
\end{center}
\vspace{-1em}
\caption{\textit{Lower}: Scaling behaviors of the quantization tokenizer when varying the size of each codebook. \textit{Upper}: Comparison of performances between RS modules of 13M and 21M parameters. Increasing the size of each codebook does not overcome the scaling saturation. The shadow regions around the lines indicate error bars from three repeated runs.}
\label{fig:codebook size scaling}
\end{figure*}

\textbf{Results.} 
We investigate how performance is affected by scaling the quantization tokenizer, varying both the number of codebooks in \cref{fig:codebook scaling} and their sizes in \cref{fig:codebook size scaling}.
This analysis is driven by two central questions: (1) \textit{Can performance gains be achieved by scaling the tokenizer in isolation?} (2) \textit{Does jointly scaling the tokenizer with the RS yield further improvements?}

Accordingly, we plot performance against both the number and size of codebooks, while also comparing RS modules of different sizes (13M and 21M parameters).
The results show that the framework achieves optimal performance when the number of codebooks is 3 and the size of each codebook is 256.
Beyond these settings, enlarging either dimension generally leads to performance degradation.
This indicates that any marginal gains from larger tokenizers are outweighed by the increased learning difficulty for the sequential recommender associated with longer and more complex SIDs.
Moreover, the performance comparison reveals no significant advantage of the 21M RS module over the 13M module, suggesting that scaling the tokenizer does not effectively overcome the saturation of the sequential recommender. In summary, we draw the following observation:

\textbf{\textit{Observation 3}: The quantization tokenizer exhibits scaling behaviors at small scales~(i.e., 3$\times$256) but further scaling will result in little gains or even performance drops~($\gamma_2 \approx 0$ in \cref{eq:codewords_scaling_law}).
}

\subsection{Are SIDs the Bottleneck? An Ablation Study}\label{subsec:ablation on codewords+RS}

From the experiments above, we observe that the scaling of the SID-based GR model exhibits a fundamental bottleneck -- neither scaling the LLM encoder nor enlarging the quantization tokenizer improves the quality of the SIDs, and little to no scaling behavior emerges.
\textbf{\textit{This suggests the bottleneck may not lie in the model's individual components but in the architectural design of the SID-based GR itself. }}
In this subsection, we ask: \emph{what will happen if we bypass SIDs and directly input semantic information into the RS module?}
We hypothesize that the SID itself could constitute a key bottleneck for scaling the entire framework, as distilling dense LLM embeddings into discrete SIDs discards substantial semantic information, preventing the RS from fully exploiting the LLM’s knowledge.

To validate this hypothesis, we conduct an ablation study by introducing external knowledge sources of collaborative filtering and semantic information.
Specifically, we inject collaborative filtering (CF) embeddings and LLM embeddings into the RS module respectively, and observe the resulting performance changes.
For CF embeddings, we train a SASRec model~\citep{SASRec} on the same training set for each dataset and use its learned item embeddings as CF embeddings.
The CF embeddings can be viewed as a knowledge source of user behaviors since the SASRec only accesses the user interaction sequences.
For LLM embeddings, we directly employ the original Flan-T5-xxl encoder outputs as item embeddings.
The LLM embeddings can be viewed as a knowledge source of item contents since the LLM encoder only accesses the item content descriptions.

We integrate these external embeddings into the RS module as follows: for the SID of item $i$, after passing through the RS embedding table, we obtain an item embedding $e_i$.
The corresponding CF or LLM embedding is then processed through a trainable MLP adapter and added to $e_i$.
The subsequent training procedure is identical to that of the standard RS module in TIGER.
For more details of the experiments, please refer to \cref{app:experiment details}.

\begin{figure}[t]
\begin{center}
\includegraphics[width=0.3\textwidth]{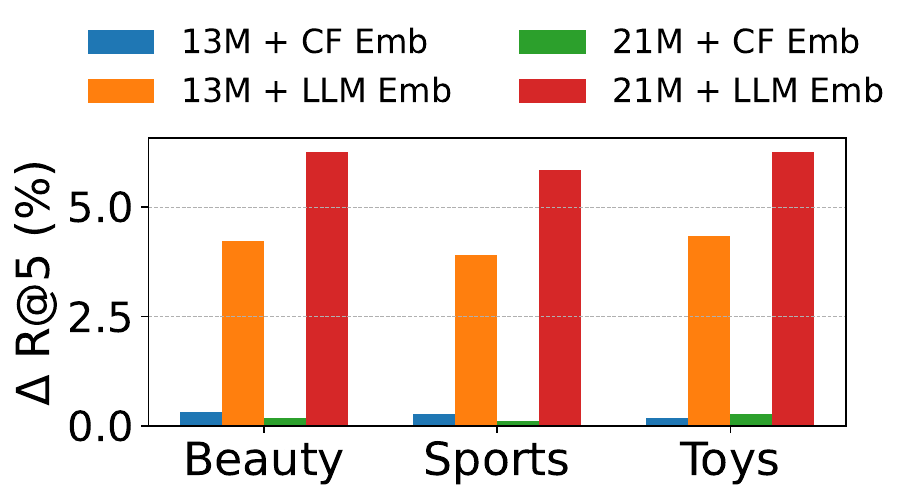}
\end{center}
\vspace{-1em}
\caption{Recall@5 gains from CF/LLM embeddings. Injecting LLM embeddings to RS module alleviates the scaling bottleneck.}
\label{fig:codebook ablation}
\end{figure}

\textbf{Results.} The results are presented in \cref{fig:codebook ablation}, where we evaluate RS modules with 13M and 21M parameters.
We find that injecting additional CF embeddings has little effect on performance, suggesting that the framework already captures collaborative filtering signals well and the injected embeddings provide largely redundant information.
In contrast, injecting LLM embeddings directly into the RS module yields substantial performance improvements.
Moreover, the gains are more pronounced for the 21M RS module than for the 13M variant, indicating that the saturation of scaling is alleviated when richer semantic information is available.
Overall, these results demonstrate that the scaling bottleneck arises from the limited semantic information encoded in SIDs. Scaling the encoder or tokenizer fails to produce more informative SIDs, underscoring a fundamental limitation of scaling SID-based GR models.
We therefore summarize the following observation:

\textbf{\textit{Observation 4}: The poor scaling of the SID-based GR lies in architectural design of the SID-based GR, constrained by its limited ability to extract semantic information using SIDs.
}


\section{Beyond the Limitations of SIDs: Scaling Up Text-based GR Model} \label{sec:scaling_LLM_RS}


We have demonstrated in the previous section that SID-based GR suffers from intrinsic scaling bottlenecks, primarily due to its sub-optimal ability to effectively leverage the rich open-world knowledge encoded in LLMs. 
Hence, in this section, we turn to an alternative paradigm that directly employs LLMs as recommender~(Text-based GR), to examine whether it can overcome the scaling bottlenecks.
We adopt the vanilla form of LLM recommender~(\cref{fig:paradigms}). 
The inputs are plain texts, consisting of simple prompts that describe the task and a list of items' text descriptions representing the user interaction sequence. 
Unlike some previous works~\citep{LLaRA,kim2025lost}, \textbf{\textit{our prompts do not contain candidate sets of the next item to ensure a fair comparison}}.
During the training stage, the model is fine-tuned through the LoRA algorithm~\citep{LoRA}. 
In inference, constrained beam search is used to generate the textual titles of next item. 

As shown in \cref{fig:paradigms}, the parameters of the model consist of two parts: the frozen LLM weights~($N_{\text{LLM}}$) and the trainable LoRA weights~($N_{\text{LoRA}}$). 
We set $N_{\text{SI}}=N_{\text{LoRA}}+\gamma N_{\text{LLM}}$ and $N_{\text{CF}}=N_{\text{LoRA}}+\beta N_{\text{LLM}}$, where $0\leq \gamma,\beta \leq 1$ are the effective parameter coefficients of the frozen LLM to learn semantic information and collaborative filtering signals, respectively. Hence, \cref{eq:empirical_scaling_law} can be re-written as \cref{eq:llm_scaling_law}:
\begin{figure}[h]

\begin{equation}
\label{eq:llm_scaling_law}
\mathrm{Recall@}k
    = R_0  -
    \eqnmarkbox[BurntOrange]{item}{\frac{A}{(N_{\text{LoRA}}+\gamma N_{\text{LLM}})^a}}
    -
    \eqnmarkbox[Emerald]{user}{\frac{B}{(N_{\text{LoRA}}+\beta N_{\text{LLM}})^{b}}}.
\end{equation}
\annotate[yshift=1em]{above,left}{item}{SI Error}
\annotate[yshift=1em]{above,right}{user}{CF Error}
\vspace{-1.5em}
\end{figure}

\begin{figure}[t]
\begin{center}
\includegraphics[width=0.5\textwidth]{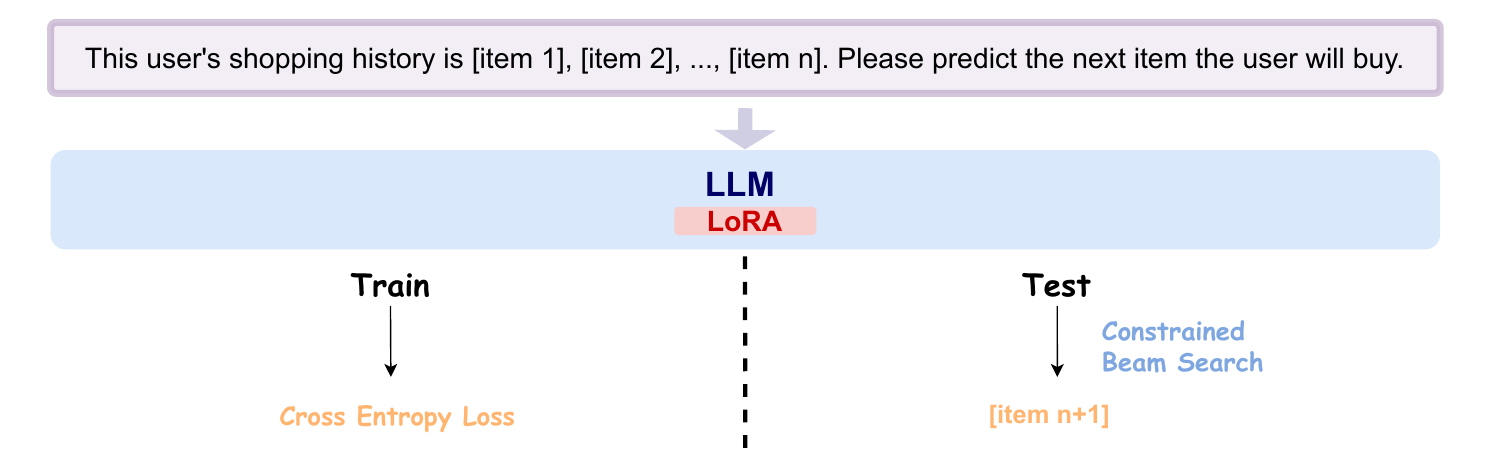}
\end{center}
\vspace{-1em}
\caption{The structure and prompts of Text-based GR model.}
\label{fig:LLM-RS basic struct}
\vspace{-1em}
\end{figure}

\begin{figure*}[t]
\begin{center}

\includegraphics[width=0.9\textwidth]{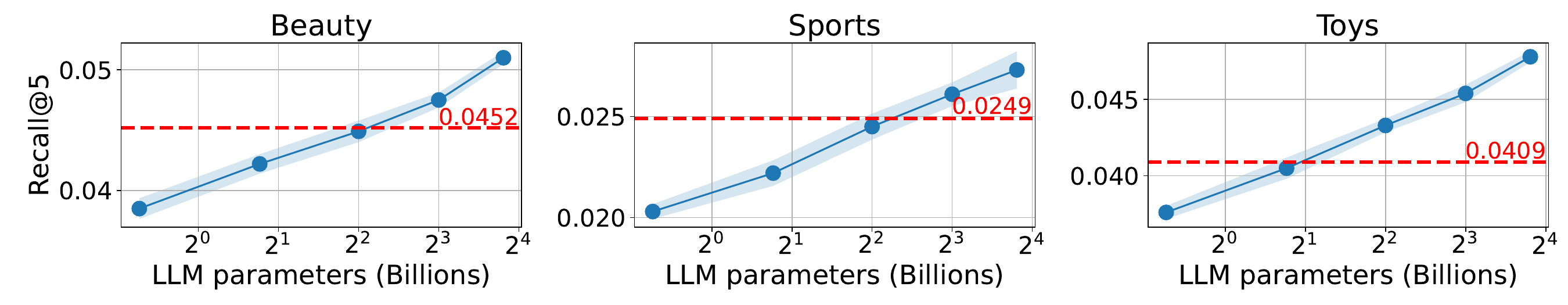}
\end{center}

\caption{The general scaling behaviors of the Text-based GR model~($N_{\text{LoRA}}\approx0.01N_{\text{LLM}}$). The \textcolor{red}{Red} dashed lines are the best scaling performance of the SID-based model. The shadow around the lines indicates error bars from three repeated runs.}
\label{fig:lora total scaling}
\end{figure*}

\begin{figure*}
\begin{center}
\includegraphics[width=0.9\textwidth]{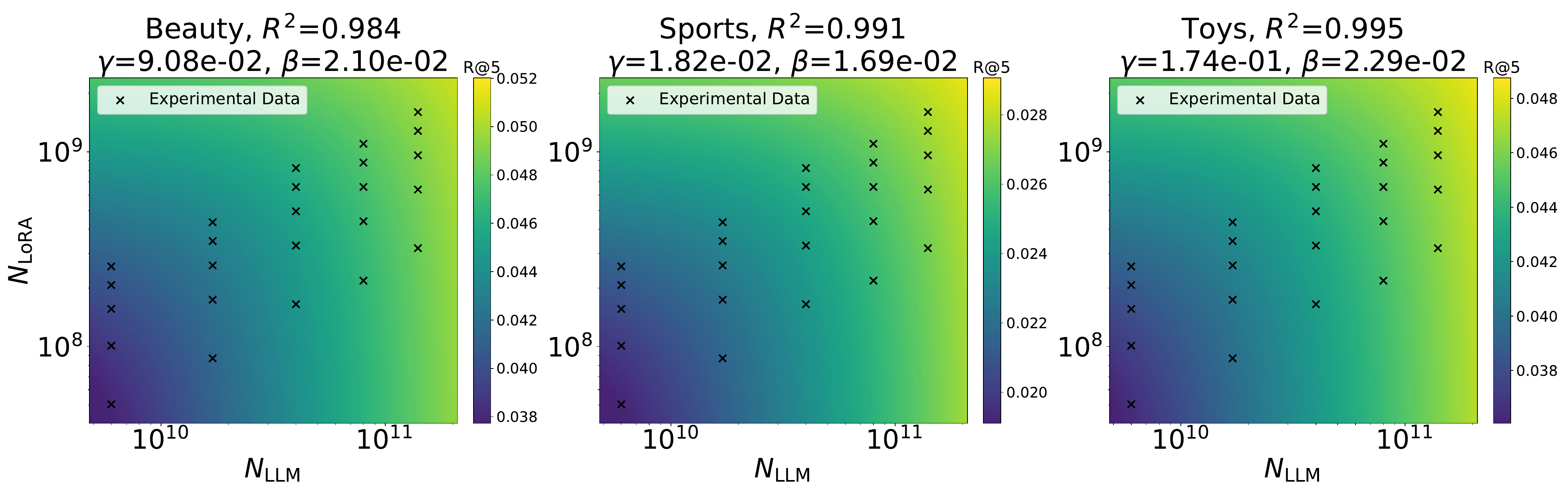}
\end{center}
\caption{Fitting \cref{eq:llm_scaling_law} to the empirical data yields high $R^2$ values, indicating strong goodness of fit. $\gamma, \beta>0$ and that the frozen LLM contributes to the learning of both CF and SI.}
\label{fig:Lora fit scaling law}
\end{figure*}

In Figure~\ref{fig:LLM-RS basic struct}, we present the framework of the Text-based GR model used in our experiments. The prompts are deliberately simple, consisting of only two sentences. The raw textual descriptions of items are organized into the user interaction sequence. We limit the maximum user interaction sequence length to be 20. 
During training, we optimize the LoRA weights~\citep{LoRA} using cross-entropy loss. 
The LoRA weights are applied on modules including 
['k\_proj', 'v\_proj', 'q\_proj', 'o\_proj', 'gate\_proj', 'up\_proj', 'down\_proj'].
For inference, we employ constrained beam search to guide the model in generating the next-item prediction. Specifically, we construct a trie from all item titles in the dataset to ensure that the model outputs only valid item titles.

In the following, we first demonstrate that the Text-based GR paradigm exhibits superior scaling properties compared to the SID-based GR~(\cref{subsec:llm_rs_scale}).
Furthermore, we show that scaling up the LLM in Text-based GR setup enhances its ability to model collaborative filtering signal, challenging the prevailing view of this paradigm (\cref{subsec:llm_rs_scale} and \cref{subsec:llm_w_cf}).

\subsection{Investigating the Scaling Behaviors of Text-based GR}\label{subsec:llm_rs_scale}

This subsection investigates the scaling properties of Text-based GR by comparing its performance to that of a SID-based GR under the same data budget. 
Specifically, we simultaneously scale both the trainable and frozen weights and plot the resulting scaling curves. 
Our experiments use the Qwen3 model series, with sizes ranging from 0.6B to 14B parameters. Throughout these experiments, the trainable LoRA weights are maintained at approximately 1\% of the total LLM parameters.

As shown in \cref{fig:lora total scaling}, simple Text-based GR models can substantially outperform SID-based GR (highlighted in \textcolor{red}{Red}) when scaled up.
Moreover, the performance consistently improves as the model scales, with no signs of saturation within the test dataset size range, suggesting that the Text-based GR exhibits stronger scaling behaviors than the SID-based GR.

Furthermore, we analyze the respective contributions of the LoRA and LLM weights to scaling.
In particular, we aim to address the question: does scaling the frozen LLM weights enhance the model’s ability to learn both user behaviors and item contents?
To provide a quantitative answer, we fit \cref{eq:llm_scaling_law} and estimate the empirical values of $\gamma$ and $\beta$.
Specifically, we vary the LoRA rank within $\{8, 16, 24, 32, 40\}$ and the LLM size within $\{0.6\text{B}, 1.7\text{B}, 4\text{B}, 8\text{B}, 14\text{B}\}$.
This yields $5 \times 5 = 25$ data points covering different combinations of LoRA and LLM sizes.
Following prior work~\citep{hoffmann2022training}, we employ Huber loss~\citep{huberloss} with the L-BFGS optimizer~\citep{LBFGS} to estimate the parameters $(R_0, A, B, \gamma, \beta, a, b)$ as:
\begin{equation}\label{eq:huber}
    \min_{R_0, A, B, \gamma, \beta, a, b}\sum_{\text{Runs } i } \text{Huber}_{\sigma=0.03} \left[ \text{log}\hat{\mathcal{R}}(N_{\text{LLM}},N_{\text{LoRA}}) - \mathcal{R}_i \right],
\end{equation}
where $\hat{\mathcal{R}}$ and $\mathcal{R}$ are the predicted and experimental values for Recall@k, respectively.
We show the fitting results in \cref{fig:Lora fit scaling law}, revealing two key findings. 
First, the high  R-square~(coefficient of determination) values indicate a strong fit of the equation to the data, confirming the effectiveness of \cref{eq:llm_scaling_law}. 
Second, both the LoRA weights and the frozen LLM demonstrate clear scaling behaviors, as evidenced by their positive scaling exponents (i.e., both $\gamma$ and $\beta$ are larger than 0). Further details on the fitting results are available in \cref{app:scaling_law_eq4}.


\textbf{\textit{Observation 5:} In the Text-based GR paradigm, both the trainable LoRA weights and frozen LLM weights show scaling behaviors. Moreover, the frozen LLM has scaling behaviors on learning both semantic information and collaborative filtering signals ($\gamma, \beta > 0$ in \cref{eq:llm_scaling_law}).}

\begin{figure*}[t]
\begin{center}

\includegraphics[width=0.9\textwidth]{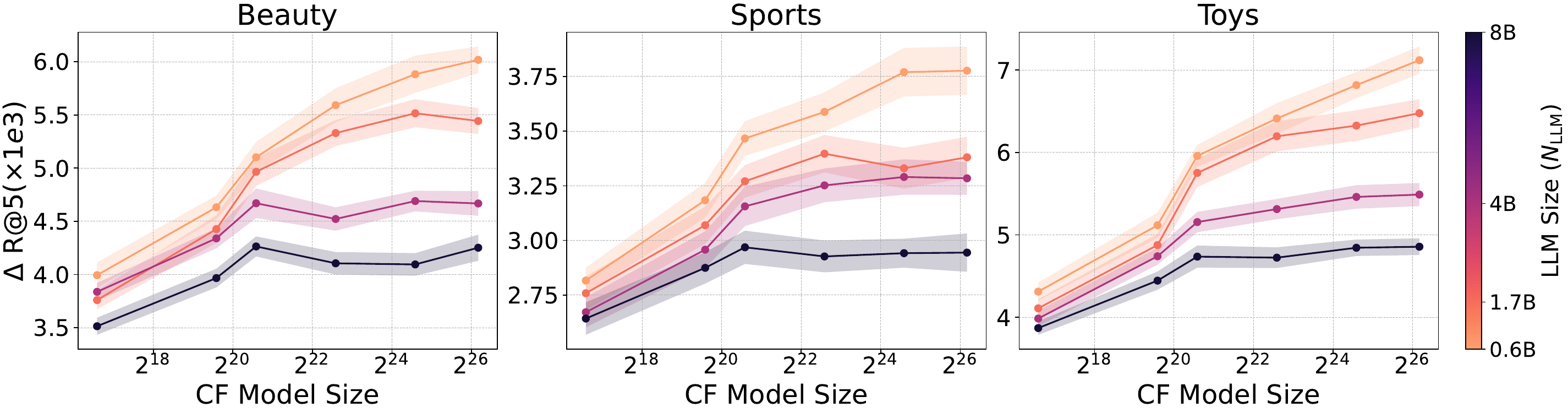}
\end{center}
\caption{Scaling behaviors of injecting external CF embeddings. The y-axis metric~($\Delta \text{Recall}@5$) measures the performance differences between models with or without external CF embeddings. For a fixed CF model size, the performance gains contributed by its CF embeddings decrease as the LLM backbone size increases, indicating that Text-based GR exhibits scaling behaviors for CF signals. The shadow around the lines indicate error bars from three repeated runs.}
\label{fig:Lora with Sasrec 1}

\end{figure*}

\subsection{Can Text-based GR Exhibit Scaling Behaviors for CF Signals?}\label{subsec:llm_w_cf}

Moreover, we further check the model scaling behaviors of Text-based GR on learning collaborative filtering signals. 
To achieve this, we incorporate external CF embeddings, a common technique for integrating such information into LLM-based recommender systems~\citep{A-LLMRec,LLaRA,kim2025lost}. 
Similar to the methodology in \cref{subsec:ablation on codewords+RS}, we use item embeddings from pre-trained SASRec~\citep{SASRec} models as the source of these CF signals. 
These embeddings are first processed by a trainable MLP adapter and then concatenated with the corresponding item's token embeddings within the LLM (see \cref{fig:LLM-RS concat_CF}). 
We conduct two sets of experiments to analyze scaling trends: first, by varying the size of the SASRec model that generates the CF embeddings, and second, by varying the LLM size ($N_{\text{LLM}}$). To ensure a fair comparison, we keep $N_{\text{LoRA}}$ fixed at $\sim$50M for all experiments. Experimental details are available in \cref{app:experiment details}.


\textbf{Results.} The scaling curves are shown in \cref{fig:Lora with Sasrec 1} with evidence supporting that $\beta$ in  \cref{eq:llm_scaling_law} is larger than 0.
We further prove this by contradiction.
Assuming $\beta=0$, then \cref{eq:llm_scaling_law} becomes
\begin{equation}
\label{eq:llm_wo_cf}
\text{Recall@}k
    = R_0  -
    \frac{A}{(N_{\text{LoRA}}+\gamma N_{\text{LLM}})^a}
    -
    \frac{B}{N_{\text{LoRA}}^b}.
\end{equation}
After adding the external CF embeddings and adapter, the scaling equation becomes
\begin{equation}
\label{eq:llm_w_cf}
\text{New Recall@}k
    = R_0  -
    \frac{A}{(N_{\text{LoRA}}+\gamma N_{\text{LLM}})^a}
    -
    \frac{B}{(N_{\text{LoRA}}+ N_{\text{SA}})^b},
\end{equation}
where $N_{\text{SA}}$ is the total number of parameters of the SASRec model and the adapter.
Hence we can get the scaling equation for the performance gains by \cref{eq:llm_w_cf} minus \cref{eq:llm_wo_cf}, written as
\begin{equation}
\label{eq:llm_CF_perf_gain}
\Delta \text{Recall@}k
    = \frac{B}{N_{\mathrm{LoRA}}^b}
    -
    \frac{B}{(N_{\text{LoRA}}+ N_{\text{SA}})^b}.
\end{equation}
Since \cref{eq:llm_CF_perf_gain} does not involve $N_{\text{LLM}}$ and we fix $N_{\text{LoRA}}$, the scaling curves with different $N_{\text{LLM}}$ should overlap.
However, the actual curves clearly diverge, indicating that the assumption is invalid and $\beta \neq 0$.
Furthermore, we observe that larger SASRec models generally yield larger gains, consistent with prior findings~\citep{SASRec_scaling}.
Yet, the performance gains of a fixed-size SASRec model decrease as the LLM backbone scales up.
This qualitatively demonstrates that increasing the frozen LLM size already enhances the model’s capacity to capture CF signals, implying $\beta > 0$.

\textbf{\textit{Observation 6:} Gains from external CF embeddings decrease as LLM backbone scales up, proving that $\beta>0$ in \cref{eq:llm_scaling_law} and Text-based GR exhibits scaling behaviors for CF signals.}

\section{Computation Costs of the Two Paradigms}\label{app:efficiency}

\begin{figure}[t]
    \centering
        \centering        \includegraphics[width=0.35\textwidth]{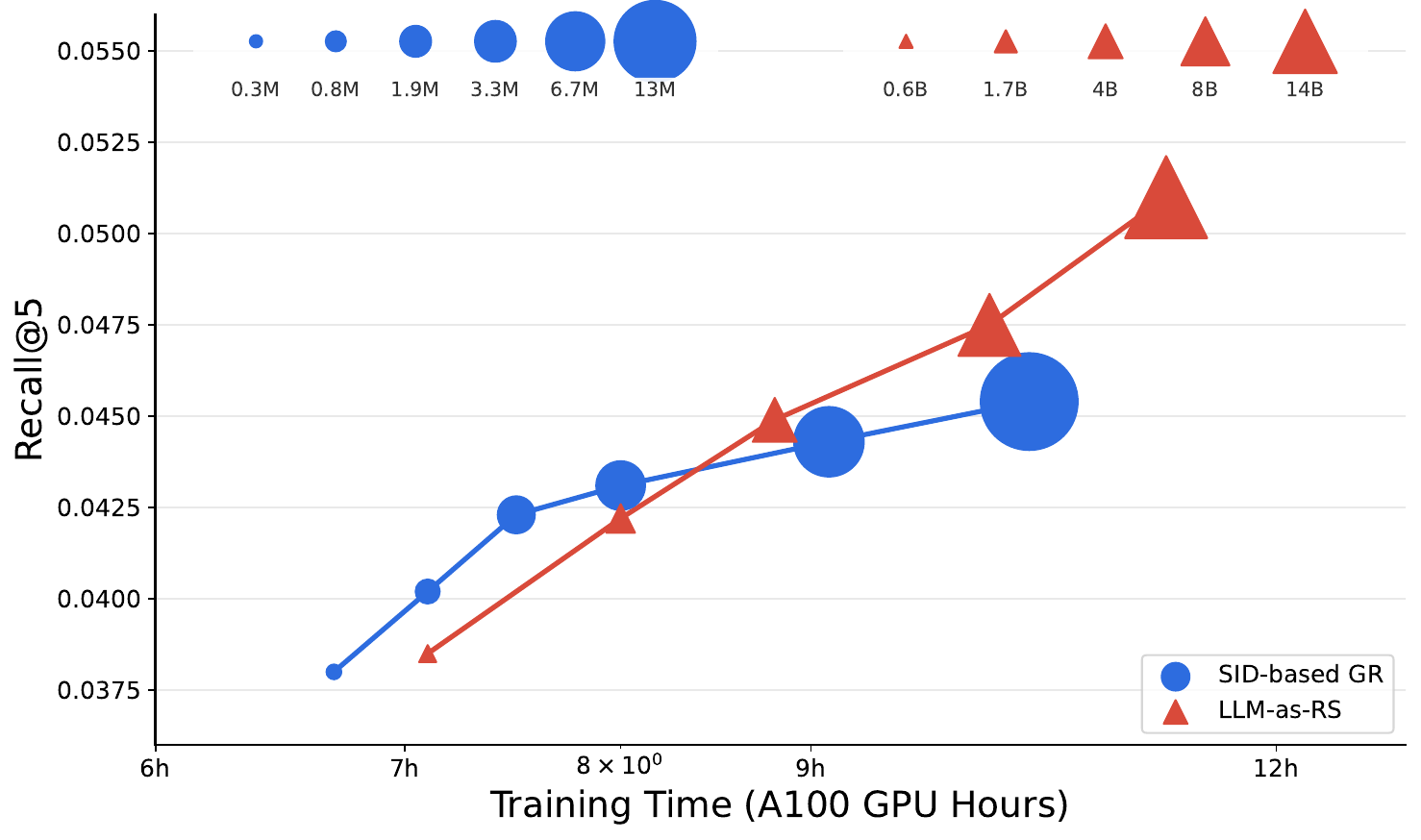}
        \vspace{-1em}
        \caption{The training time and performance of models of different sizes. Training time is measured by the GPU hours required for a model to be trained to converge.}
        \label{fig:first}
        \vspace{-1em}
    \end{figure}

\begin{figure}[t]
        \centering
    \includegraphics[width=0.35\textwidth]{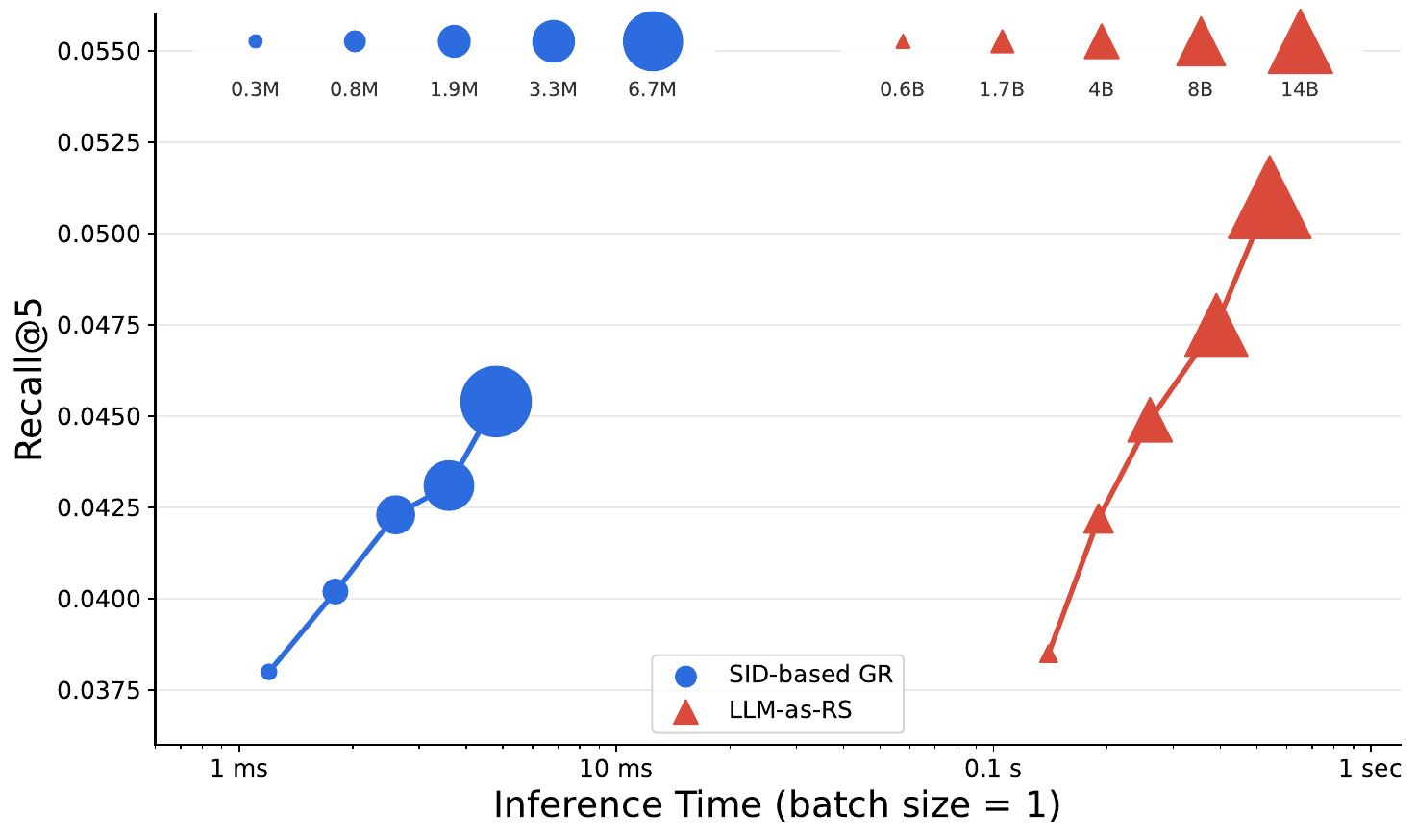}
    \vspace{-1em}
        \caption{The inference time and performance of models of different sizes.}
        \label{fig:second}
        \vspace{-1em}
\end{figure}

We compare the training and inference efficiency of the two paradigms on the Beauty dataset, as shown in \cref{fig:first} and \cref{fig:second}. Training time is measured by the GPU hours required for a model to converge. We observe that under limited training time or computational budgets, the SID-based method performs better. However, as models scale, the performance of Text-based GR surpasses that of SID-based GR with comparable training time. On the other hand, inference costs for SID-based GR are substantially lower than those of Text-based GR, primarily because SID-based GR generates only four IDs per item, whereas Text-based GR must generate full titles, typically longer than four tokens. Thus, SID-based GR maintains efficiency advantages when resources are constrained, while Text-based GR becomes preferable when budgets allow and performance is the primary objective.


\section{Cold-Start Experiments}\label{app:cold start}

In this section, we compare the cold-start ability of the two paradigms. For each dataset, we randomly select 500 items as cold-start items in the test set and remove them from the corresponding training sets. We then train the models and evaluate their Recall@10 on these cold-start items. The results, presented in \cref{tab:cold_start}, show that Text-based GR consistently outperforms SID-based GR, indicating stronger transferability to unseen items.

\begin{table}[h]
\centering
\caption{The comparison on the cold-start items. Text-based GR consistently outperforms SID-based GR.}
\begin{tabular}{c|cc}
\hline
       & SID-based GR   & Text-based GR      \\ \hline
Beauty  & 0.018 & \textbf{0.030} \\ 
Sports  & 0.006 & \textbf{0.014}   \\ 
Toys  & 0.022 & \textbf{0.028}  \\ \hline
\end{tabular}
\label{tab:cold_start}
\end{table}
\section{Conclusions and Discussions}\label{sec:Discussions amd Conlusions}

In this work, we examined the scaling behaviors of current mainstream paradigms in generative recommendation. 
Our work is the first to show that SID-based GR has an essential bottleneck in utilizing semantic information, limiting the scaling of the whole framework with model capacity.
In contrast, the Text-based GR paradigm can better learn both collaborative filtering signals and semantic information, making it a more favorable candidate for recommendation foundation models.
Nonetheless, Text-based GR currently suffers from efficiency limitations, as discussed in \cref{app:efficiency}. Specifically, SID-based GR remains preferable when budgets are constrained and efficiency is critical, whereas Text-based GR is more suitable when performance is prioritized and resources are sufficient.
Overall, our findings suggest that  how to overcome the shortcomings of current paradigms and fully unlock the potential of LLMs is a critical topic  for further research.

\textbf{Limitations.}
\textit{(a) Modality coverage of item content.}
Our experiments represent items solely by \emph{textual} descriptions during encoding and generation. Real-world item contents could be multimodal (e.g., images for products, frames and audio for videos, or speech/music for podcasts). Future work can extend the analysis to multi-modal data. \textit{(b) Scope of results.}
Our scaling observations focus on mainstream instantiations of current paradigms (SID-based GR with residual quantization over text encodings; Text-based GR with textual inputs and outputs). Other models might exhibit different scaling behaviors. Results were obtained on open-source academic datasets. Switching the datasets to the industrial level might affect the model scaling behaviors as well.


\bibliographystyle{ACM-Reference-Format}
\bibliography{main}

\appendix
\section{Dataset Details}\label{app:dataset_details}

Here we present the details of the datasets we use in the experiments. We mainly use three datasets from the Amazon Reviews Datasets~\citep{BLAIR}. We list the basic statistics of the datasets in ~\cref{tab:dataset_stats}.

\begin{table}[h]
\centering
\caption{Statistics of the datasets used in our experiments.}
\vspace{-1em}
\begin{tabular}{lrrr}
\toprule
\textbf{Dataset} & \textbf{\# users} & \textbf{\# items} & \textbf{\# actions}  \\
\midrule
\textit{Beauty} & 22,363 & 12,101 & 198,502 \\
\textit{Toys and Games} & 19,412 & 11,924 & 167,597 \\
\textit{Sports and Outdoors} & 35,598 & 18,357 & 296,337\\\bottomrule
\end{tabular}
\label{tab:dataset_stats}
\end{table}

\section{Experiment Details}\label{app:experiment details}

\subsection{General Settings}
We first present the general experimental settings, which are shared across all experiments. Detailed settings specific to each experiment are provided in the corresponding sections. All models are optimized using AdamW~\citep{AdamW}. For each training process, we search the learning rate over \{1e-2, 1e-3, 1e-4\} and select the best-performing value. All experiments are conducted on 8$\times$80G and 8$\times$40G NVIDIA Tesla A100 GPUs.

\subsection{Experiment Settings of \cref{sec:scaling_codewords}}\label{subsec:app_experiment_sid}

\textbf{Quantization tokenizer structure.}
A uniform codebook configuration notation is \((L,W)\) (e.g. \((3,256)\) or $3\times256$ meaning \(L=3\) layers each with \(W=256\) entries); for the nonuniform case one writes the vector of sizes \((W^{(0)},W^{(1)},\dots,W^{(L-1)})\). 
\textbf{We use the number of the codebooks and the size of each codebook for $L$ and $W$ in the main text, respectively.}

\textbf{Scaling RS modules} For all the experiments with the SID-based GR model, we keep the length of input user interaction sequence at 20.
For the scaling experiments in \cref{subsec:code_RS_scaling}, we enlarge the RS module according to \cref{tab:scale RS}.

\begin{figure}[h!]
\begin{center}
\includegraphics[width=0.5\textwidth]{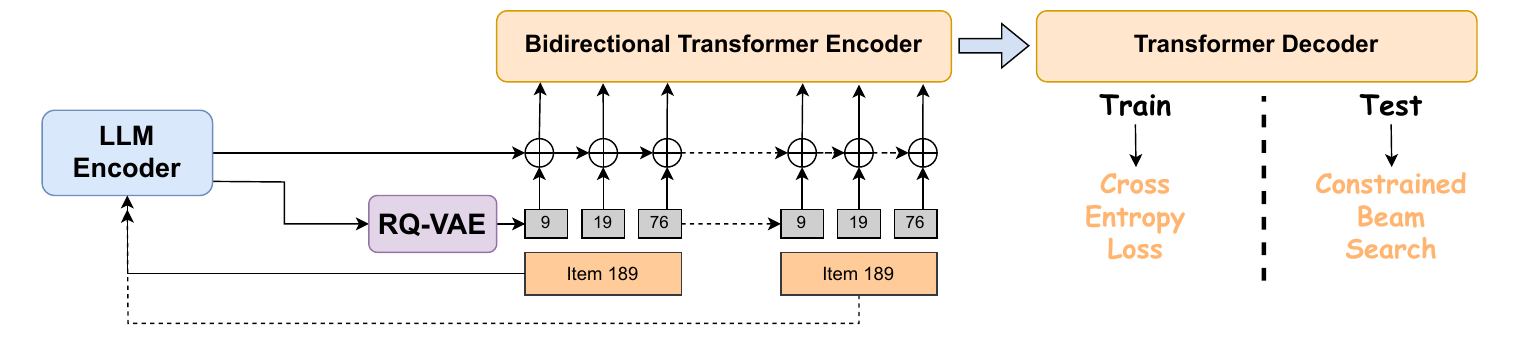}
\end{center}
\vspace{-1em}
\caption{The illustration of injecting the LLM embeddings into SID-based GR model.}
\label{fig:SID_add_SI}
\vspace{-1em}
\end{figure}

\begin{figure}[h!]
\begin{center}
\includegraphics[width=0.5\textwidth]{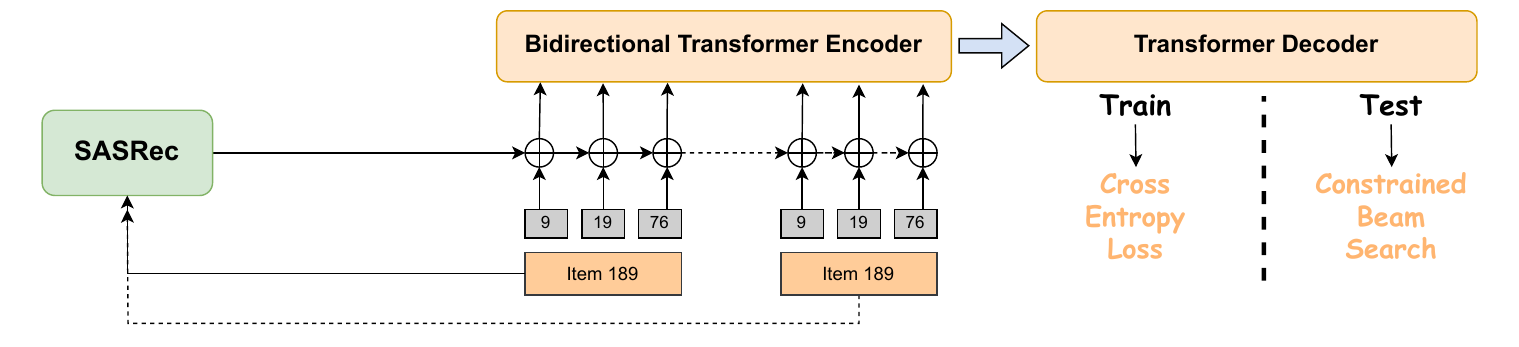}
\end{center}
\vspace{-1em}
\caption{The illustration of injecting the CF embeddings into SID-based GR model.}
\label{fig:SID_add_CF}
\vspace{-1em}
\end{figure}

\textbf{Injecting CF/LLM embeddings} For the experiments in \cref{subsec:ablation on codewords+RS}, we fetch and inject the LLM embeddings as shown in \cref{fig:SID_add_SI}, and produce and inject the CF embeddings as shown in \cref{fig:SID_add_CF}.

\begin{table}[h!]
\centering
\caption{The details of scaling RS module in SID-based GR. The \#Params are just rounded values. The \#Layers are the same for both encoder and decoder in the module.}
\vspace{-1em}

\begin{tabular}{r|ccccc}
\hline
\textbf{\#Params} & \textbf{\#Layers} & $\mathbf{d_{\text{model}}}$ & \textbf{\#Heads} & $\mathbf{d_{\text{kv}}}$ & $\mathbf{d_{\text{ff}}}$ \\
\hline
336{,}000 & 1 & 64  & 3  & 64 & 512  \\
778{,}000 & 2 & 64  & 3  & 64 & 512  \\
1{,}900{,}000 & 5 & 64  & 3  & 64 & 512  \\
3{,}300{,}000 & 9 & 64  & 3  & 64 & 512  \\
6{,}700{,}000 & 3 & 128 & 6  & 64 & 1024 \\
13{,}000{,}000 & 4 & 128 & 6 & 64 & 1024 \\
21{,}000{,}000 & 7 & 128 & 6  & 64 & 1024 \\
43{,}000{,}000 & 8 & 192 & 9  & 64 & 1536 \\
88{,}000{,}000 & 9 & 320 & 15 & 64 & 2560 \\
192{,}000{,}000 & 20 & 384 & 18 & 64 & 3072 \\
\hline
\end{tabular}
\label{tab:scale RS}
\end{table}

\subsection{Experiment Settings of \cref{sec:scaling_LLM_RS}}

\begin{figure}[h!]
\begin{center}
\includegraphics[width=0.5\textwidth]{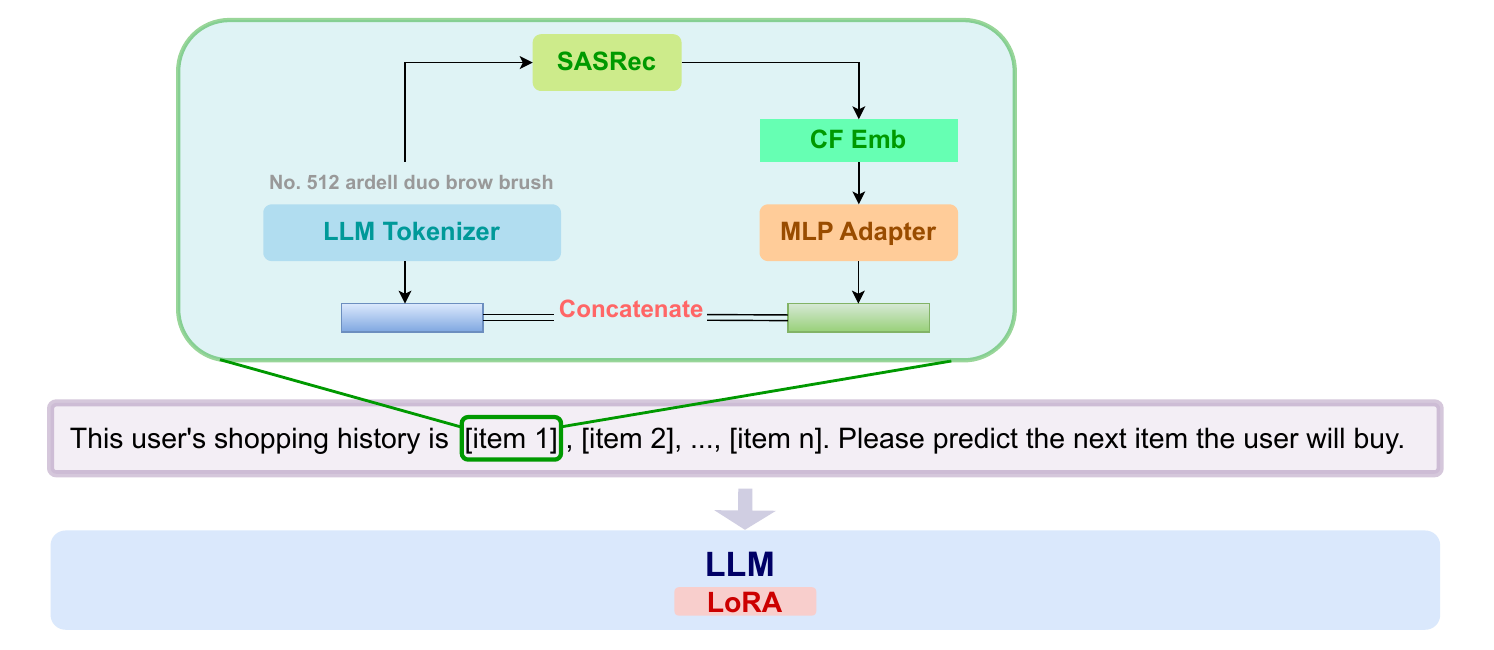}
\end{center}
\vspace{-1em}
\caption{The illustration of injecting the CF embeddings into Text-based GR model by concatenation.}
\label{fig:LLM-RS concat_CF}
\vspace{-1em}
\end{figure}

For experiments in \cref{subsec:llm_w_cf}, we produce and inject the external CF embeddings as shown in \cref{fig:LLM-RS concat_CF}. And we scale up SASRec models according to \cref{tab:details of scaling sasrec}.

\begin{table}[h]
\centering
\caption{The details of scaling the SASRec model.}
\vspace{-1em}
\begin{tabular}{rccc}
\hline
\textbf{\#Params} & \textbf{\#Layers} & $\mathbf{d_{\text{model}}}$ & \textbf{\#Heads} \\
\hline
   98,304      & 2  & 64   & 2  \\
  786,432      & 4  & 128  & 4  \\
1,572,864      & 8  & 128  & 4  \\
6,291,456      & 8 & 256  & 8  \\
25,165,824     & 8 & 512  & 8  \\
75,497,472    & 24 & 512 & 8 \\
\hline
\end{tabular}
\label{tab:details of scaling sasrec}
\end{table}

\section{Fitting Equation~\ref{eq:codewords_scaling_law} to Empirical Data}\label{app:equation3_fit}

In the main text, we have shown that $\gamma_1$ and $\gamma_2$ in \cref{eq:codewords_scaling_law} equal zero. Hence, \cref{eq:codewords_scaling_law} can be re-written as

\begin{equation}
\mathrm{Recall@}k
    = R_0  -
    \frac{A}{N_{\text{RS}}^a}
    -
    \frac{B}{N_{\text{RS}}^{b}}
\end{equation}

Then we will fit the equation to the empirical data in \cref{subsec:code_RS_scaling}, following the same process as \cref{subsec:llm_rs_scale} with \cref{eq:huber_eq_sid}. The results are shown in \cref{tab:fitted_params_SID}. 
The R-square values are larger than 0.9, indicating the equation fits real data well.

\begin{equation}\label{eq:huber_eq_sid}
    \min_{R_0, A, B, a, b}\sum_{\text{Runs } i } \text{Huber}_{\sigma=0.03} \left[ \text{log}\hat{\mathcal{R}}(N_{\text{RS}}) - \mathcal{R}_i \right]
\end{equation}

\renewcommand{\arraystretch}{1.5}
\begin{table}[h!]
\centering
\caption{Fitted parameters of scaling laws of SID-based GR.}
\vspace{-1em}
\begin{tabular}{c|ccc}
\hline
       & Beauty & Sports & Toys \\ \hline
$\text{R-square}$  & 0.94  & 0.97   & 0.94 \\
$R_0$  & 0.4529   & 3e-1   & 1.7e-1 \\
$A$    & 16.8  & 24.8 & 6.1 \\
$B$    & 1e-2 & 1e-2 & 1e-2 \\
$a$    & 0.6 & 0.63 & 0.52 \\
$b$    & 2.23 & 1.97 & 2.02 \\ \hline
\end{tabular}

\label{tab:fitted_params_SID}
\end{table}

\section{More Results of the Scaling Law Fitting}\label{app:scaling_law_eq4}

\renewcommand{\arraystretch}{1.5}
\begin{table}[h!]
\centering
\caption{The fitted empirical parameters in Section~\ref{subsec:llm_rs_scale}.}
\vspace{-1em}
\begin{tabular}{c|ccc}
\hline
       & Beauty & Sports & Toys \\ \hline
$R_0$  & 3e-1    & 3e-1   & 1.7e-1 \\
$A$    & 9.9e2  & 9.87e2 & 2.19e-1 \\
$B$    & 3.4e-1 & 3.35e-1 & 9.93e2 \\
$\gamma$ & 9.08e-2 & 1.82e-2 & 1.74e-1 \\
$\beta$  & 2.10e-2 & 1.69e-2 & 2.29e-2 \\
$a$    & 1.98e1 & 2.02e1 & 2.47e-2 \\
$b$    & 1.39e-2 & 9.47e-3 & 2.02e1 \\ \hline
\end{tabular}
\label{tab:fitted_params}
\end{table}

\begin{table}[h!]
\centering
\caption{Held-out fitting errors (the lower the better) of Equation~\ref{eq:llm_scaling_law} when $\beta=0$ and $\beta>0$.}
\vspace{-1em}
\begin{tabular}{c|cc}
\hline
 & $\beta=0$ & $\beta>0$ \\ \hline
Beauty & 4.1e-4$\pm$2e-5 & \textbf{3.2e-4} $\pm$1e-5\\ 
Sports & 1.1e-3 $\pm$3e-5& \textbf{3.4e-4}$\pm$2e-5 \\ 
Toys & 1.2e-3$\pm$4e-5 & \textbf{9.1e-4}$\pm$2e-5 \\ \hline
\end{tabular}
\label{tab:held_errors}
\end{table}

In Section~\ref{subsec:llm_rs_scale}, we fit Equation~\ref{eq:llm_scaling_law} with the empirical data.
We only report the values of $\gamma$ and $\beta$, here we report the values of all the parameters in Table~\ref{tab:fitted_params}.
Moreover, we compare the held-out errors of Equation~\ref{eq:llm_scaling_law} and Equation~\ref{eq:llm_wo_cf} to give further evidence that $\beta>0$. 
Specifically, we randomly choose 20\% of the experimental data in Figure~\ref{fig:Lora fit scaling law} as test data, and use the remaining data to fit the two equations.
Then we will use the two fitted equations to predict the values of test data and calculate the held-out errors as \cite{finetune_scaling}. The errors are listed in Table~\ref{tab:held_errors}.
We find that $\beta>0$ will lead to smaller held-out errors, which justifies our choice for Equation~\ref{eq:llm_scaling_law} and further supports our argument that $\beta>0$.
Although our scaling equations achieve a good fit, they also carry the risk of overparameterization. For example, in \cref{tab:adjusted R-square}, we compare the \textbf{adjusted $R^2$} of our equation against that of the single-term equation. The advantage of our equation shrinks under this stricter metric, suggesting that greater caution is warranted when generalizing the equation to new scenarios.

\begin{table}[h]
\centering
\caption{Adjusted $R^2$ Comparison.} \label{tab:adjusted R-square}
\vspace{-1em}
\begin{tabular}{lccc}
\toprule
 & \textbf{Beauty} & \textbf{Sports} & \textbf{Toys} \\
\midrule
Single-term & 0.905 & 0.942 & \textbf{0.921} \\
SI/CF decomposition & \textbf{0.922} & \textbf{0.959} & 0.918 \\
\bottomrule
\end{tabular}
\end{table}

\section{An Initial Study on Data Scaling}

\begin{table}[h]
\centering
\caption{Performance comparison between SID and Text across different sample sizes.}\label{tab:data scale}
\vspace{-1em}
\begin{tabular}{lcccccc}
\toprule
\textbf{\#Samples} & \textbf{2e5} & \textbf{4e5} & \textbf{6e5} & \textbf{8e5} & \textbf{1e6}  & \textbf{2e6} \\
\midrule
SID  & 0.0314 & 0.0360 & 0.0395 & 0.0419 & 0.0430  & 0.0443 \\
Text & 0.0321 & 0.0370 & 0.0404 & 0.0432 & 0.0445  & 0.0472 \\
\bottomrule
\end{tabular}
\label{tab:samples}
\end{table}

\noindent As an exploration on data scaling, we provide results on Amazon-M2~\cite{Amazon-M2}, an industrial-scale dataset that is approximately 50 times larger than datasets used in the main text. Specifically, we fix the model size at 4B parameters and gradually increase the training data size. As in \cref{tab:data scale}, we observe that text-based methods consistently outperform SID-based one, and that the performance gap gradually widens as the amount of training data increases.

\end{document}